\documentclass{article}

\usepackage[dblblindworkshop, final]{neurips_2026}
\workshoptitle{AI for Science}

\makeatletter
\renewcommand{\@noticestring}{%
  Accepted at the AI for Science workshop (NeurIPS 2026).%
}
\makeatother

\usepackage[utf8]{inputenc}
\usepackage[T1]{fontenc}
\usepackage{hyperref}
\hypersetup{
  pdftitle    = {The Price of an Explanation: What Demanding Verifiable Artifacts Costs, and What It Buys Back},
  pdfauthor   = {Edgar A. Duéñez-Guzmán, Eduardo Dueñez, Edgar Said Hernandez Sanchez, Suzanne Sadedin},
  pdfsubject  = {AI for Science workshop, NeurIPS 2026},
  pdfkeywords = {verifiable artifacts, program synthesis, domain-specific language, evolutionary search, ARC-AGI-2, quantum error correction},
}
\usepackage{url}
\usepackage{booktabs}
\usepackage{amsfonts}
\usepackage{amsmath}
\usepackage{nicefrac}
\usepackage{microtype}
\usepackage{xcolor}
\usepackage{graphicx}

\newcommand{\SepSplit}{\texttt{s20260904}}
\newcommand{\SepCThreeLerPct}{4.18\%}
\newcommand{\SepCThreeMs}{5.66}
\newcommand{\SepCThreeTimeXPTwoD}{0.64}
\newcommand{\SepCFourLerPct}{4.14\%}
\newcommand{\SepCFourMs}{20.6}
\newcommand{\SepCSevenMs}{6.15}
\newcommand{\SepCEightMs}{6.96}
\newcommand{\SepCNineMs}{5.42}
\newcommand{\SepCTenLerPct}{4.07\%}
\newcommand{\SepCTenMs}{29.0}
\newcommand{\SepAcLerPct}{3.57\%}
\newcommand{\SepAcMs}{31.8}
\newcommand{\SepPTwoDLerPct}{5.45\%}
\newcommand{\SepPTwoDMs}{8.82}
\newcommand{\SepLdpcOsdZeroLerPct}{4.39\%}
\newcommand{\SepLdpcOsdZeroMs}{903.8}
\newcommand{\SepRelayFiveLerPct}{3.17\%}
\newcommand{\SepRelayFiveMs}{190.0}
\newcommand{\SepRelayFiveTimeXPTwoD}{21.6}
\newcommand{\SepRelayNineLerPct}{2.80\%}
\newcommand{\SepRelayNineMs}{270.6}
\newcommand{\SepRelayNineTimeXPTwoD}{30.7}
\newcommand{\SepCThreeZVsPTwoD}{+31.8}
\newcommand{\SepCThreeErrCutVsPTwoD}{23\%}
\newcommand{\SepFront}{seven}
\newcommand{\KnobDampedLerPct}{5.15\%}
\newcommand{\KnobDampedShots}{4\,096}
\newcommand{\KnobUndampedLerPct}{47.66\%}
\newcommand{\KnobUndampedShots}{512}
\newcommand{\KnobLerRatio}{9.3}
\newcommand{\KnobEdgesRatio}{1.04}
\newcommand{\KnobLiftedLerPct}{47.4\%}
\newcommand{\KnobLiftedShots}{1\,206}
\newcommand{\OctDampSplit}{\texttt{s20261004}}
\newcommand{\OctDampShots}{200\,000}
\newcommand{\OctDampPTwoMs}{365.08}
\newcommand{\OctDampPTwoDMs}{8.91}
\newcommand{\OctDampPThreeMs}{372.35}
\newcommand{\OctDampPThreeDMs}{21.71}
\newcommand{\OctDampOsdZeroSpeedup}{41.0}
\newcommand{\OctDampOsdZeroLerRatio}{1.27}
\newcommand{\OctDampOsdZeroZ}{+25.8}
\newcommand{\OctDampOsdCsFrontSpeedup}{17.1}
\newcommand{\OctDampOsdCsFrontLerRatio}{1.10}
\newcommand{\OctDampOsdCsFrontZ}{+9.2}
\newcommand{\OctAdjSplit}{\texttt{s20261004}}
\newcommand{\OctAdjShots}{200\,000}
\newcommand{\OctAdjRefusals}{zero}
\newcommand{\OctAdjProxyOverTimeMin}{1.01}
\newcommand{\OctAdjProxyOverTimeMax}{1.26}
\newcommand{\OctAdjMachine}{Intel Xeon Platinum 8581C}
\newcommand{\OctAdjCThreeCostXPTwoD}{0.58}
\newcommand{\OctAdjCThreeTimeXPTwoD}{0.565}
\newcommand{\OctAdjCThreeZVsPTen}{\ensuremath{-}19.6}
\newcommand{\OctAdjCThreeZVsPTwoD}{+33.4}
\newcommand{\OctAdjCFourCostXPTwoD}{2.0}
\newcommand{\OctAdjCFourTimeXPTwoD}{1.882}
\newcommand{\OctAdjCFourZVsPTen}{\ensuremath{-}20.6}
\newcommand{\OctAdjCFourZVsPTwoD}{+35.8}
\newcommand{\OctAdjCSevenCostXPTwoD}{0.46}
\newcommand{\OctAdjCSevenTimeXPTwoD}{0.456}
\newcommand{\OctAdjCSevenZVsPTen}{\ensuremath{-}31.9}
\newcommand{\OctAdjCSevenZVsPTwoD}{+23.3}
\newcommand{\OctAdjCEightCostXPTwoD}{0.57}
\newcommand{\OctAdjCEightTimeXPTwoD}{0.534}
\newcommand{\OctAdjCEightZVsPTen}{\ensuremath{-}19.6}
\newcommand{\OctAdjCEightZVsPTwoD}{+33.1}
\newcommand{\OctAdjCNineCostXPTwoD}{0.54}
\newcommand{\OctAdjCNineTimeXPTwoD}{0.537}
\newcommand{\OctAdjCNineZVsPTen}{\ensuremath{-}22.5}
\newcommand{\OctAdjCNineZVsPTwoD}{+30.8}
\newcommand{\OctAdjCTenCostXPTwoD}{2.7}
\newcommand{\OctAdjCTenTimeXPTwoD}{2.389}
\newcommand{\OctAdjCTenZVsPTen}{\ensuremath{-}22.0}
\newcommand{\OctAdjCTenZVsPTwoD}{+35.9}
\newcommand{\OctAdjETwentySevenCostXPTwoD}{2.4}
\newcommand{\OctAdjETwentySevenTimeXPTwoD}{2.145}
\newcommand{\OctAdjETwentySevenZVsPTen}{+9.4}
\newcommand{\OctAdjETwentySevenZVsPTwoD}{+54.7}
\newcommand{\OctAdjPTenCostXPTwoD}{3.0}
\newcommand{\OctAdjPTenTimeXPTwoD}{2.67}
\newcommand{\OctAdjPTwoCostXPTwoD}{25.1}
\newcommand{\OctAdjPTwoDCostXPTwoD}{1.0}
\newcommand{\OctAdjPTwoDTimeXPTwoD}{1.00}
\newcommand{\OctAdjPThreeCostXPTwoD}{26.0}
\newcommand{\OctAdjPFiveCostXPTwoD}{40.0}
\newcommand{\OctAdjPFiveTimeXPTwoD}{35.57}
\newcommand{\OctRelayFiveMs}{172.8}
\newcommand{\OctRelayFiveTimeXPTwoD}{19.7}
\newcommand{\OctRelayNineMs}{272.0}
\newcommand{\OctRelayNineTimeXPTwoD}{31.0}
\newcommand{\OctRelayPTwoDMs}{8.74}
\newcommand{\NovConfirmShots}{32\,768}
\newcommand{\NovCThreeArm}{seeded, work proxy}
\newcommand{\NovCFourArm}{unseeded, work proxy}
\newcommand{\NovCSevenArm}{seeded, wall-clock}
\newcommand{\NovCEightArm}{seeded, wall-clock}
\newcommand{\NovCNineArm}{seeded, wall-clock}
\newcommand{\NovCTenArm}{unseeded, wall-clock}
\newcommand{\NovETwentySevenArm}{seeded, work proxy}
\newcommand{\NovETwentySevenDistPruned}{0}
\newcommand{\NovETwentySevenVerdictFull}{retuning of BP+AC}
\newcommand{\HypDslOps}{38}
\newcommand{\HypWorkCap}{60}
\newcommand{\HypBudgetSweeps}{2\,000}
\newcommand{\HypBudgetOpCalls}{256}
\newcommand{\HypBudgetCandidates}{10\,100}
\newcommand{\HypBudgetColumnOps}{982\,256}
\newcommand{\HypCpusPerActor}{2}
\newcommand{\HypDriftPNinetyFiveCpuOne}{51 \%}
\newcommand{\HypFlagRateCpuOne}{30 \%}
\newcommand{\HypDriftMedianCpuTwo}{0.9 \%}
\newcommand{\HypFlagRateCpuTwo}{0 \%}
\newcommand{\CostCounters}{seven}
\newcommand{\CostWeightEdgesMinsum}{1.00}
\newcommand{\CostWeightEdgesProdsum}{1.91}
\newcommand{\CostWeightGfTwoWords}{0.28}
\newcommand{\CostWeightCandWords}{2.25}
\newcommand{\CostWeightScanElems}{0.30}
\newcommand{\CostWeightOpCalls}{0.00}
\newcommand{\CostWeightClastTicks}{0.00}
\newcommand{\CostMinsumOverProdsum}{0.52}
\newcommand{\CostFitPNinetyFive}{11.8\%}
\newcommand{\CostFitHoldoutPNinetyFive}{21.2\%}
\newcommand{\CostFitWidestCone}{22.3\%}
\newcommand{\CostFitRSq}{0.998}
\newcommand{\CostFitPrograms}{252}
\newcommand{\CostSelectionMargin}{1.4}
\newcommand{\CostMarginFloor}{1.31}
\newcommand{\CostRef}{12\,924.58}
\newcommand{\CostRankCFourTau}{0.973}
\newcommand{\CostRankCFourSpearman}{0.998}
\newcommand{\CostRankCFourPrograms}{252}
\newcommand{\CostRankMsukTau}{0.935}
\newcommand{\CostRankMsukSpearman}{0.992}
\newcommand{\XbPtwoLer}{5.53}
\newcommand{\XbPtwoVsPtenZ}{-48.9}
\newcommand{\XbPtenLer}{3.44}
\newcommand{\XbPtenVsPtwoZ}{+48.9}
\newcommand{\XbPfiveLer}{2.86}
\newcommand{\XbPfiveVsPtwoZ}{+58.0}
\newcommand{\XbPfiveVsPtenZ}{+15.5}
\newcommand{\XbPtwoLdpcLer}{4.35}
\newcommand{\XbPtwoLdpcVsPtwoZ}{+25.8}
\newcommand{\XbPtwoLdpcVsPtenZ}{-25.7}
\newcommand{\XbPthreeLdpcLer}{3.84}
\newcommand{\XbPthreeLdpcVsPtwoZ}{+36.7}
\newcommand{\XbPthreeLdpcVsPtenZ}{-11.5}
\newcommand{\XbEeightythreeLer}{3.17}
\newcommand{\XbEeightythreeErrors}{6\,347}
\newcommand{\XbEeightythreeMs}{36.2}
\newcommand{\XbEeightythreeVsPtwoZ}{+53.8}
\newcommand{\XbEeightythreeVsPtenZ}{+7.5}
\newcommand{\XbEeightythreeVsPfiveZ}{-8.8}
\newcommand{\XbEeightythreeVsRelayZ}{-9.8}
\newcommand{\XbEeightythreeVsAeDslZ}{+28.7}
\newcommand{\XbEeightythreeTimeOverAeDsl}{0.69}

\newcommand{\XbAeDslLer}{4.26}
\newcommand{\XbAeDslMs}{52.9}
\newcommand{\XbAeDslVsPtwoZ}{+30.2}
\newcommand{\XbAeDslVsPtenZ}{-24.7}
\newcommand{\XbRelayLer}{2.82}
\newcommand{\XbRelayErrors}{5\,635}
\newcommand{\XbRelayMs}{270.6}
\newcommand{\XbRelayVsPtwoZ}{+58.6}
\newcommand{\XbRelayVsPtenZ}{+16.7}
\newcommand{\XbRelaySfiveLer}{3.17}
\newcommand{\XbRelaySfiveVsPtwoZ}{+51.3}
\newcommand{\XbRelaySfiveVsPtenZ}{+7.0}
\newcommand{\XbAeFreeLer}{1.95}
\newcommand{\XbAeFreeErrors}{3\,904}

\newcommand{\XbAeFreeMs}{168}

\renewcommand{\SepSplit}{A}
\renewcommand{\OctAdjSplit}{B}
\renewcommand{\OctDampSplit}{B}

\title{The Price of an Explanation: What Demanding\\
       Verifiable Artifacts Costs, and What It Buys Back}

\author{%
  Edgar A. Duéñez-Guzmán$^{1}$
  \quad Eduardo Dueñez$^{1,2}$ \\[2pt]
  \textbf{Edgar Said Hernandez Sanchez}$^{1}$
  \quad \textbf{Suzanne Sadedin}$^{1}$ \\[2pt]
  $^{1}$Gibran.AI
  \quad $^{2}$The University of Texas at San Antonio \\[2pt]
  \texttt{\{duenez, eduardo, said, suzanne\}@gibran.ai}
}

\begin{document}

\maketitle

\begin{abstract}
Generative models produce many candidate solutions but provide no intrinsic signal by
which to rank them. One mitigation is to demand a \emph{verifiable artifact}: a
program that can be executed against the available evidence. A common concern is that
requiring executable verification incurs costs in both solution quality and computational budget.
We measure these two costs on the $120$-task ARC-AGI-2 evaluation set. Demanding a
verifiable artifact from an LLM reduced the score from $93$ to $68$ of $120$ tasks ($77\%$ to
$57\%$). Of those $25$ tasks, $21$ were reached by no program-producing configuration we ran,
at any budget, by sampling or by search; yet for $27$ of the $28$ tasks that only the
answer-only arm solved, some program arm held a program that reproduced every demonstration
and still failed the test.
The bottleneck is obtaining an explanation that \emph{generalizes}, not finding one. A
short evolutionary search over two already-sampled programs, at no additional model cost,
beats best-of-$2$ sampled code by $9.2 \pm 1.4$ tasks, and over a single sampled program
beats $1$-shot sampling by $9.9 \pm 1.3$. Moreover, because a program can be checked against
the training examples, any program-producing protocol can terminate early once a task is
explained: under the same stopping rule, best-of-$n$ sampling declines up to $38\%$ of its
budget and the evolved arm up to $49\%$, whereas an answer-only baseline cannot terminate
early. Applying the same methodology to decoder synthesis for quantum error correction yields
interpretable decoders evaluated against a certified panel of published decoders on held-out
exam shots.
\end{abstract}

\section{Introduction}

Large language models (LLMs) are increasingly finding applications in science~\citep{
taylor2022galactica, boiko2023autonomous, gil2014amplify}. Their ability to
ingest and synthesize vast amounts of information makes them an appealing tool
for hypothesis generation, literature review, experimental design, and
more~\citep{wang2023scientific}. However, substantial risks arise when incorporating
LLMs into the scientific process~\citep{fecher2025friend, birhane2023science},
including confidently claiming plausible-sounding but ultimately wrong results~\citep{
rodionov2026traces, simhi2025trust}. A common mitigation is
to demand a verifiable artifact from the LLM in the form of code, even
when the problem is not intrinsically a coding one~\citep{liang2023code, novikov2025alphaevolve,
ying2024internlm, hubert2026olympiad}. The code acts not only as an answer, but as a
verifiable \emph{explanation} of where the answer comes from. However, demanding
verifiability incurs a cost, for example in the form of additional inference-time
computation~\citep{snell2024scaling} or repeated sampling~\citep{brown2024large}.

Program synthesis driven by a language model and steered by an evaluator has
produced results in mathematics~\citep{novikov2025alphaevolve},
algorithms~\citep{romeraparedes2024funsearch} and robotics~\citep{liang2023code},
among others. Self-consistency and best-of-$n$ sampling~\citep{wang2023selfconsistency}
are the established baselines for any such loop. On ARC specifically, sampling programs
from a language model and filtering them on the demonstrations is the
method of~\citet{greenblatt2024arc} and of hypothesis search~\citep{wang2024hypothesis};
BARC~\citep{li2024barc} trains the generator on synthetic tasks and combines program
induction with direct output prediction, and HySynth~\citep{barke2024hysynth} uses the
model as a learned prior over a grammar that guides enumerative synthesis. Improving
model-written code by evolutionary variation is the subject of evolution through large
models~\citep{lehman2022elm}, language-model crossover~\citep{meyerson2023lmx} and
LLM-guided genetic improvement~\citep{brownlee2023gi}. While these works investigate how
to maximize program-synthesis accuracy, they do not isolate what requiring an executable
program costs relative to a matched answer-only baseline on the same model, nor quantify
how much inference budget verification recovers via early termination; providing those
controlled measurements is the primary contribution of this paper.

\begin{figure}[t]
  \centering
  \includegraphics{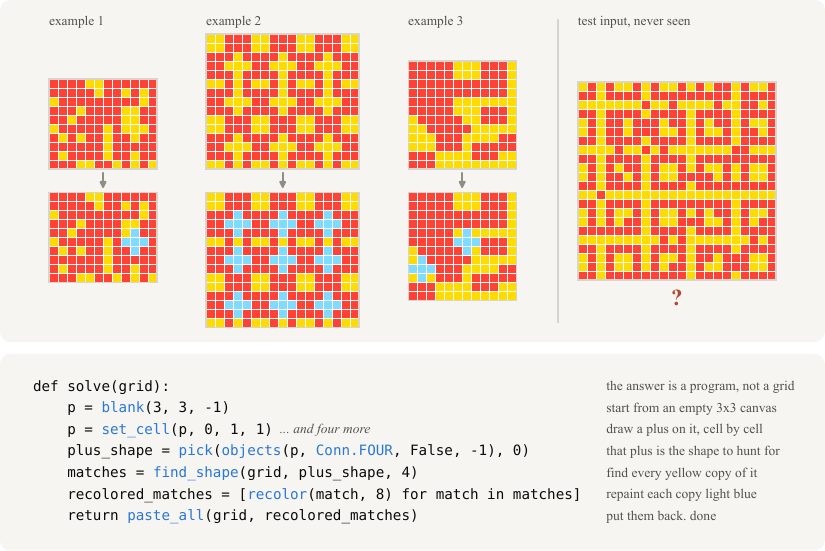}
  \caption{\textbf{A verifiable artifact.} Task
    \texttt{1818057f} of the public ARC-AGI-2 evaluation set: three worked
    examples, an unseen input, and the program our search returned. The rule,
    which is to recolor every plus-shaped yellow cluster light blue, is stated
    nowhere and must be inferred, and what is submitted is the \emph{program}
    rather than the grid. The program is written in the restricted vocabulary
    of a domain-specific language (DSL; highlighted in blue; \texttt{4} and
    \texttt{8} are ARC's color codes for yellow and light blue). It reproduces
    all three worked examples; a program with that property is what we call an
    explanation, and its cost is what this paper measures. On
    the unseen input it reproduces the held-out answer exactly. Four
    other \texttt{set\_cell} lines are elided where marked.}
  \label{fig:arc-example}
\end{figure}

Code, unlike free-form natural language, admits verification by formal proof, which LLMs can
increasingly generate~\citep{first2023baldur} at token budgets in the hundreds of
millions~\citep{lazaropoulos2026foundational}, or more cheaply by generating and verifying
together~\citep{li2026p}. Verifier-guided search over code in Dafny and
Coq~\citep{brandfonbrener2024vermcts} and neurosymbolic self-debugging for mathematical
code~\citep{nezhad2026symcode} both report more accurate code at lower cost; in both, the problem
space admits no non-verifiable answer, so no cost comparison with an answer-only baseline is
possible. Here we consider problems whose solutions do not \emph{require} a verifiable artifact,
but where one \emph{could} be demanded, on the Abstraction and Reasoning Corpus,
ARC-AGI-2~\citep{chollet2019arc, chollet2025arcagi2}. Each ARC task consists of
\emph{train} input-output pairs exemplifying rules that transform an input grid
into an output grid, which must then be applied to held-out \emph{test} inputs.
The benchmark is purposefully small ($1{,}000$ training, $120$ evaluation tasks)
to resist purely data-driven approaches in favor of abstraction and reasoning.

Producing the correct output grids is sufficient to succeed in ARC. One may,
however, additionally demand that the solution be given as code that maps the
inputs to the desired outputs (Figure~\ref{fig:arc-example}). That is, one can
demand a verifiable \emph{explanation}. A successful strategy to achieve high scores
in the ARC-AGI-1 version of the benchmark was
to create domain-specific languages (DSLs) and search that space, sometimes aided
by a language model~\citep{hodel2024addressing, chollet2024arc}. However,
the most successful strategies for ARC-AGI-2 use DSLs more incidentally,
to generate large datasets of ARC-like tasks, thus shifting the problem space back
to where data-driven approaches perform well~\citep{chollet2026arcprize}. This
strategy allowed language models to dominate the ARC-AGI-2 Kaggle competition.

Here, we first quantify the costs of verifiability by asking one leading model to codify its
solutions using a DSL derived exclusively from the ARC-AGI-2 training tasks. Second, we explore how
a brief, effectively free search process on generated code---making no further model calls---can
mitigate, and in some cases negate, these costs. We used a simple evolutionary algorithm for this search, though other search
procedures (e.g., Monte Carlo tree search~\citep{swiechowski2023monte}) could be substituted.
Finally, we transfer the methodology to a real-world domain, quantum error
correction~\citep{gottesman1997stabilizer}, where we synthesize a new \emph{decoder}
(error-correcting algorithm) of the belief-propagation family~\citep{pearl2014probabilistic}
with ordered-statistics post-processing (BP+OSD)~\citep{panteleev2021degenerate,
roffe2020decoding} for bivariate-bicycle codes~\citep{bravyi2024high}.

\section{Methods}
\label{sec:setup}

The methodology we evaluate composes a language model with evolutionary
search: the model proposes candidate programs, scored on the train pairs, filtered and
ranked. The search uses steady-state grammatical
evolution~\citep{oneill2003grammatical} (MAP-Elites~\citep{mouret2015illuminating} in
\S\ref{sec:sound} and \S\ref{sec:qec}), and our LLM-only baselines are best-of-$B$ computed from the first $B$ samples of one ordered cache rather than from
independent re-runs. We compare three experimental arms (Figure~\ref{fig:spend}):
\begin{description}
    \item[Answer-only (A0)] Direct LLM prediction of output grids, without an executable explanation.
    \item[Explainable (X)] Independent LLM program sampling steered to use the DSL.
    \item[Evolved (explainable) (E)] LLM-seeded DSL programs refined via evolutionary search and optional in-loop LLM calls.
\end{description}

Nothing about an answer-only (A0) candidate can be checked before it is
scored. An explainable arm (X and E) instead produces candidate \texttt{solve}
functions, executes each against the training pairs, and scores it by the
fraction of correct outputs. Train-perfection---reproducing all of the
train outputs---is what makes a candidate an \emph{explanation}, a verifiable
property of the artifact whose soundness we measure in \S\ref{sec:sound}.
Two properties of a program artifact should be distinguished: verifiability comes from
execution against the demonstrations and belongs to any program, in a DSL or in free
Python (\S\ref{sec:dsl_cost}), whereas the DSL governs what a \emph{search} can do with the
artifact and what a reader can audit as a bounded vocabulary.

\paragraph{Submission rule, shared by X and E.} ARC allows two attempts per test input.
Both program arms submit the same way: candidates are ranked by train score (ties broken in
favor of the later sample), attempt~1 is the top candidate, and attempt~2 is the highest-ranked
candidate whose test predictions differ from attempt~1's, a diversity hedge that
inspects only the test \emph{inputs}. At $B{=}2$ the explainable arm holds two samples and
submits both whenever their predictions differ, so the train-score ranking cannot
change what it submits; the paired comparison of \S\ref{sec:convert} therefore does
not depend on the ranking rule. Its contribution for $B>2$ is measured in
App.~\ref{sec:adaptive}.

We use \texttt{gemini-3.5-flash} (medium thinking budget) across all arms as a capable
reasoning model that does not saturate the benchmark ($<95\%$ accuracy). The search
never sees test outputs, and an evolved run stops once it holds two train-perfect programs
with distinct predictions: training fitness and the candidate's own outputs on the test
inputs are the only signals a run may use. The explainable arm is reported at a fixed
best-of-$B$ throughout, and \S\ref{sec:stopping} also evaluates it under the same stopping
rule.

\paragraph{Substrate and credit rule.} All ARC results are on the full $120$-task
ARC-AGI-2 public evaluation set. ARC's protocol is best-of-$2$ per test input and $45$ of the
$120$ tasks carry more than one test input ($167$ in all). We use the scoring protocol of the
\textbf{ARC leaderboard}---score each input independently, report
the task's mean~\citep{arcprize2026benchmarking}---everywhere except Table~\ref{tab:cost}.
In that table, we use \emph{strict} credit to count which tasks have been solved (i.e., every
test input has the correct output). The two are never mixed inside a comparison.

\paragraph{The answer-only (A0) and explainable (X) arms.} A0 predicts grids directly,
taking $k{=}12$ samples and two attempts by majority vote; because it has no program to
re-run, it pays per test \emph{input} ($16.7$ calls/task at $k{=}12$). The explainable arm X
steers the LLM to write DSL code transforming train inputs into outputs; all sampling
curves read a single ordered $k{=}16$ cache per style so that best-of-$B$ for
$B \in \{1,2,4,8,16\}$ is a nested prefix of the same samples.

\paragraph{The evolved arm (E).} Given a per-task LLM call budget
$B \in \{1,2,4,8,16\}$:
(i)~sample $\min(2, B)$ seed programs using X's prompt;
(ii)~if both are train-perfect with distinct test predictions, submit both and \emph{stop};
(iii)~if one is train-perfect, lock it as attempt~1 and evolve, seeded with both, for a
hedged attempt~2;
(iv)~if neither is, evolve seeded with both and submit a hedged best-of-$2$.
Evolution spends any remaining budget on in-loop LLM calls at rate $p=0.03$ per birth while
budget lasts.

\paragraph{The DSL, error bars, and QEC transfer.} From a conceptual
analysis of the $1{,}000$ training tasks (App.~\ref{sec:dsl}), we built a $91$-function
Python DSL without inspecting the $120$ evaluation tasks (free-form Python is compared in
App.~\ref{sec:nodsl}). We report $\mathbb{E}[\text{solves}]$, using the directly measured
run-to-run spread of $1.86$ solves (App.~\ref{sec:err}) rather than within-run replica
noise ($\mathrm{SE}\ 0.5$--$1.1$) for cross-run comparisons. To test transferability, we
instantiate the same design on decoder synthesis for the $[[72,12,6]]$ bivariate-bicycle
qLDPC code~\citep{bravyi2024high} with a $\HypDslOps$-operation BP+OSD
DSL~\citep{panteleev2021degenerate,roffe2020decoding} (App.~\ref{sec:qecdetails}).

\section{Results}

\subsection{The cost of demanding a verifiable artifact}
\label{sec:cost}

We compare the answer-only arm A0 against the explainable
arms on the $120$-task evaluation set as solved-task sets rather than as rates,
so that the cost of the demand can be written as a set difference. To place all arms on
a common basis, a task is credited to an arm in Table~\ref{tab:cost} if it is solved in a
majority (at least two) of its three runs: three replicas for E, three seed draws for X,
and three runs for A0. The union of explainable arms is taken over the two program
protocols, X (best-of-$16$ sampling) and E (pooled over $B \in \{1, 2, 4, 8, 16\}$),
both under the DSL prompt.

\begin{table}[t]
  \caption{The cost of demanding a verifiable artifact, on $120$ ARC-AGI-2 evaluation
  tasks, DSL prompt. The union row is an \emph{oracle} over the explainable arms, not a
  protocol (\S\ref{sec:cost}). \textbf{This table uses strict credit rather than
  leaderboard score}, because set differences require a binary verdict per task.
  Every arm enters as the majority of its three runs.}
  \label{tab:cost}
  \centering
  \begin{tabular}{lr}
    \toprule
    & tasks solved \\
    \midrule
    A0, grid prediction (no explanation available)      & $93$ \\
    best explainable arm (E, at any budget)             & $68$ \\
    \textbf{union of explainable arms (X $\cup$ E)}     & $\mathbf{72}$ \\
    union across all arms (A0 $\cup$ X $\cup$ E)        & $100$ \\
    \midrule
    solved by A0 only (no program arm)                  & $28$ \\
    \textbf{solved only with an explanation}            & $\mathbf{7}$ \\
    \bottomrule
  \end{tabular}
\end{table}

\begin{figure}[t]
  \centering
  \includegraphics{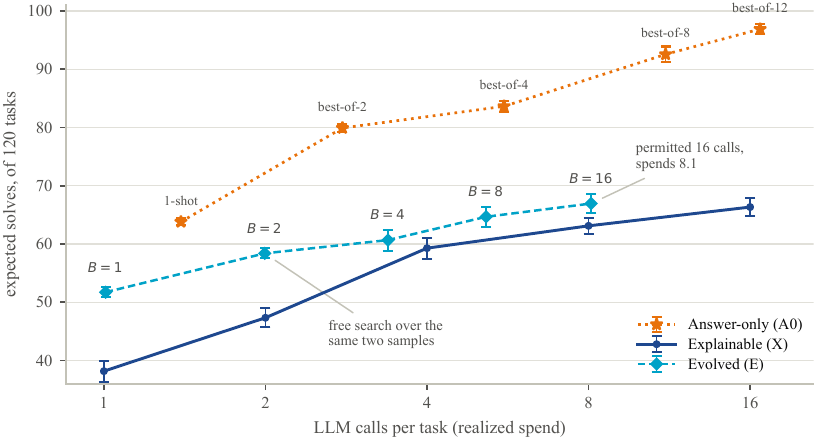}
  \caption{Solves against \emph{realized} spend. The answer-only arm
  (orange stars) predicts outputs directly. Solutions cost $1.39\times k$
  per task, because each test input requires a separate LLM call.
  The explainable arm (navy circles) is shown at fixed best-of-$n$ program sampling
  (Figure~\ref{fig:stopping} and Table~\ref{tab:adaptive} also give it the stopping rule). The
  evolved arm (cyan diamonds) has an LLM call budget $B$ used for
  seeding up to $2$ solutions for evolution, and for in-loop LLM improvements
  afterwards; it stops once it holds two distinct train-perfect candidates
  (\S\ref{sec:stopping}).
  Per-point bars are \emph{within-run} replica noise (E, 3 replicas),
  within-style draw noise (X), or block-and-run noise (A0).}
  \label{fig:spend}
\end{figure}

The cost of demanding a verifiable artifact can be measured in two ways:
(1) \textbf{The computational cost}: the A0 arm at one sample per test input ($1.39$ calls/task)
scores $63.8$ (leaderboard credit); the evolved arm reaches that level at $B{=}8$ ($64.7$,
$5.16$ calls) and the explainable arm at $B{=}16$ ($66.3$, $9.9$ calls), inside the
$1.8$--$4.2$ solve spread across seed draws.
(2) \textbf{The accuracy cost}: at the answer-only arm's largest budget ($k=12$,
$16.7$ LLM calls), the demand costs $93 - 68 = 25$ solved tasks
(Table~\ref{tab:cost}). Against the union of DSL program arms, $28$ tasks solved by A0 were
reached by \emph{no} DSL arm, at any budget, by sampling or by search, while $7$ were solved
only by a DSL arm (net $21$; the remaining $4$ is an \emph{oracle margin} over the union,
which no single protocol achieves without knowing in advance which arm to run). Adding the
free-form prompt of \S\ref{sec:dsl_cost} (fourteen program arms in all) raises the union from
$72$ to $81$, leaving $21$ tasks reached by no program configuration we ran against $9$
solved only by one---bounds over our finite configuration set and one model, not properties
of the program representation. Figure~\ref{fig:spend} shows the spend curve.

The exchange is two-way: seven tasks are solved \emph{only} by an arm that
produced an explanation, whereas no answer-only run reached them across three runs of twelve
samples per test input. Crucially, of the $28$ tasks solved only by A0, $27$ had a
train-perfect program in at least one program arm (pooled across X's three draws and E's
nine runs); thus a program explaining the demonstrations was found on nearly all of them,
and what was submitted did not generalize (\S\ref{sec:sound}, App.~\ref{sec:tasks}). The two
subsets also differ in structure: $54\%$ of A0-only tasks carry multiple test inputs versus
$26\%$ of tasks solved by both modalities, and their input grids are larger (median $456$ vs.\
$316$ cells). Because A0 receives $k{=}12$ samples per test \emph{input} ($16.7$ calls per
task, more than the $B\!\le\!16$ budgets of the program arms), the comparison is
unfavorable to the program arms on budget by construction.

\subsection{The cost of using a DSL}
\label{sec:dsl_cost}

We re-ran both program arms with a second prompt that provides no DSL and requests
pure Python with NumPy and SciPy; evolutionary search then ingests the model's code as an
abstract syntax tree, the representation grammatical evolution operates on.

Free-form code leads the DSL in the explainable arm at every budget, in $15$ of $15$
(budget $\times$ draw) cells. Under evolution it still leads at every budget, but the gap
is sign-unstable across seed draws, and evolution adds more to the DSL arm ($+9.2$
solves) than to the free-form arm ($+6.1$), so the two nearly tie at $B{=}2$
(App.~\ref{sec:nodsl}). This reflects a trade-off between model fluency and structured
search: the DSL lowers initial sampling accuracy but provides a bounded, auditable
vocabulary amenable to AST-level evolutionary operators, whereas free-form Python leverages
the model's pretraining fluency at the cost of $4.4\times$ higher native-library crash rates
(\S\ref{sec:limits}). Verifiability is identical for both.

\subsection{Search over programs already purchased}
\label{sec:convert}

Because an explanation is checkable against the training pairs without a model call, it can
drive syntactic search over already-sampled programs. We compare the evolved arm at $B = k$
against the best-of-$k$ explainable arm for $k \in \{1, 2\}$, where both arms consume the
same seed program(s) and the evolved arm spends nothing further on the model; pairing on three
independent seed draws cancels the seed-draw variance that dominates each arm separately.

Given the same model calls, the evolved arm outperforms the explainable arm by
$+9.89 \pm 1.26$ solves at one call ($+9.7, +12.2, +7.8$; paired $t$-test $p{=}0.016$) and
$+9.22 \pm 1.40$ at two ($+10.0, +6.5, +11.2$; $p{=}0.022$), each mean $\pm$ SE over three
paired seed draws (strict credit: $+8.67 \pm 0.96$ and $+7.56 \pm 1.68$; free-form:
$+7.56 \pm 1.21$ and $+6.11 \pm 1.03$; Table~\ref{tab:inloop}, Figure~\ref{fig:spend}).
Evolution is worth at least a doubling of the seed budget at no additional API cost, trading
model calls for modest wall time (mean $100$\,s, median $7$\,s per task at $B{=}2$ under a
$900$\,s deadline; App.~\ref{sec:costdetails}).

\subsection{An adaptive stopping rule}
\label{sec:stopping}

\begin{figure}[t]
  \centering
  \includegraphics{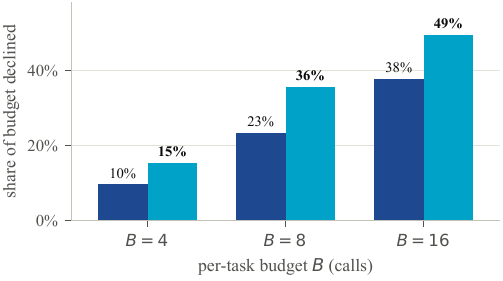}
  \caption{Savings due to the stopping rule: the share of its permitted budget each
  program arm \emph{declines}, DSL prompt, mean over three independent seed draws.
  Navy: the explainable arm given the rule---sequential sampling that stops once it
  holds two train-perfect programs with distinct predictions (App.~\ref{sec:adaptive}).
  Cyan: the evolved arm, whose remaining budget buys in-loop model calls. The answer-only
  arm, run at fixed $k$, has no checked signal on which to stop. The two arms' solves at
  these points are within $0.6$ of each other at $B{=}16$ ($66.3$ against $66.9$) and
  within the seed-draw spread everywhere (Table~\ref{tab:adaptive}).}
  \label{fig:stopping}
\end{figure}

A program that reproduces every train example is, by the protocol's rule, finished. This
property belongs to the artifact, not to the search: an arm that samples programs sequentially
can apply the same stopping rule, whereas the answer-only arm has no executable check against
the demonstrations (sample consensus~\citep{aggarwal2023adaptive} tests agreement, not
verification). We measure realized spend as a fraction of permitted
budget for both program arms at $B \in \{4, 8, 16\}$ over three seed draws, giving the
explainable arm the rule offline on the same ordered sample caches (stop once two
train-perfect programs with distinct predictions are held; App.~\ref{sec:adaptive}).

Figure~\ref{fig:stopping} shows the result (also visible in Figure~\ref{fig:spend}, where
every evolved point above $B{=}2$ sits left of its budget's tick). The sampler declines
$10\%$, $23\%$ and $38\%$ of its budget at $B = 4, 8, 16$ while scoring within $2$ solves of
fixed best-of-$B$ ($66.3$ vs.\ $66.3$ at $B{=}16$). The evolved arm declines $15\%$, $36\%$
and $49\%$ at $66.9$ solves at $B{=}16$ ($8.09$ calls/task), saving additional in-loop budget
once a task is explained. Stopping at the \emph{first} explanation saves $73\%$ at $B{=}16$
but loses ten solves ($56.0$), because the first train-perfect program is often not the right
one (\S\ref{sec:sound}). Verifiability is thus not only an audit property paid for at the end,
but a scheduling property that pays back during the run to any protocol holding a checkable
artifact.

\subsection{Soundness of the explanation, and the generalization gap}
\label{sec:sound}

An explanation is, in practice, a \emph{necessary} but not sufficient condition for
correctness. Across $1{,}079$ of $1{,}080$ trials (three search arms $\times$ three in-loop
rates $\times$ $120$ tasks; one lost to a dead worker), \emph{lucky matches}---programs
that produce the correct test output without reproducing the train pairs---number
\textbf{0--2 per cell}. Filtering by train-perfection therefore discards almost nothing it
should have kept, thus justifying the stopping rule of \S\ref{sec:stopping}.

\paragraph{An explanation of the demonstrations is often not the rule.}
\label{sec:selection}
The converse direction does not hold: \textbf{$106$ of $120$ tasks yielded a train-perfect
explanation, yet at most $68$ converted to a
solve}, so on at least $38$ of the $106$ tasks ($36\%$) the verified explanation failed to
generalize. The binding constraint was not finding a program consistent with the
demonstrations, but obtaining and selecting one that captures the intended rule. At $B{=}16$
(three draws) a train-perfect program was sampled on $102$--$106$ tasks and one also
test-correct on $71$--$72$, while the selector converted $61$--$64$ (an oracle over the
sampled pool recovers $8$--$11$ solves). Nor is it search budget: quadrupling both
search caps changed no outcome across $188$ budget-bound cells, and stopping at the first
explanation lost ten solves (\S\ref{sec:stopping}).

Consistent with this, a learned selector did not close the gap. A classifier trained on the ARC training
split to rank candidate programs (gradient-boosted and logistic models, held-out AUC
$0.95$--$0.97$) re-ranked the submission in $41\%$ of cells ($90\%$ of cells where the
selector chooses both attempts), yet changed solved tasks by less than the variance across
four control runs of the same arm ($55.5 \pm 0.19$ vs.\ $55.3$ and $55.7$). Although the
same classifier improved selection by $+4.4$ solves on pools of independent LLM samples,
features that discriminate among independent samples (prediction consensus, rank, and score
gaps) fail to separate the closely related genealogical variants produced by a search that
halts at train-perfection. Even so, an executable artifact is what makes this gap
observable and diagnosable, whereas an answer-only model conceals it entirely.

\paragraph{Evolutionary search without a language model solved nothing.} We found
no train-perfect programs across $120$ tasks and three seeded replicas, measured
on the same DSL vocabulary as every other arm reported here. The explanation's
value therefore derives from the composition of a generator that proposes with
a verifiable artifact that filters and guides, and not from the search
machinery on its own.

\subsection{Transfer: decoder synthesis for quantum error correction}
\label{sec:qec}

To assess transferability, we applied the same methodology to synthesize quantum
error-correction decoders for the $[[72,12,6]]$ bivariate-bicycle code under
circuit-level noise at $p = 0.004$ (App.~\ref{sec:qecdetails},~\citealp{evorobust2026}).
Every result is adjudicated on $200{,}000$ held-out exam shots (read once per frozen
algorithm, never fed back into optimization) against a panel of published decoders written
in the $\HypDslOps$-operation DSL and certified per shot against the authors'
implementations (\texttt{ldpc}, \texttt{relay-bp}). Inside the search, cost is measured by a
$\CostCounters$-counter primitive-level work proxy fitted to native Rust execution across
$\CostFitPrograms$ programs ($R^2 = \CostFitRSq$, Kendall $\tau = \CostRankCFourTau$ rank
agreement with wall-clock time on the fitting nodes and $\tau = \CostRankMsukTau$ on the
benchmark workstation); adjudicated cost verdicts must hold on both the work proxy (at a
$\CostSelectionMargin\times$ margin) and clean single-core wall-clock time per shot, which
agree within $\OctAdjProxyOverTimeMin$--$\OctAdjProxyOverTimeMax\times$ on every timed
finalist with $\OctAdjRefusals$ proxy verdicts refused.

\paragraph{Auditing reference implementations via the DSL.} Writing BP+OSD-0 in the
DSL exposed a message-damping parameter that the reference \texttt{ldpc} package does not
export. Undamped, the flooding min-sum front end exhibits a $\KnobUndampedLerPct$ logical
error rate (LER) versus $\KnobDampedLerPct$ when damped at $\KnobEdgesRatio\times$ the
belief-propagation work ($\KnobLerRatio\times$ fewer errors), and re-parameterizing each
\texttt{ldpc} decoder with damping yields a $\OctDampOsdCsFrontSpeedup$--$\OctDampOsdZeroSpeedup\times$
speed-up (split \OctDampSplit, \OctDampShots{} shots). Expressing reference algorithms in
the same transparent vocabulary as candidate solutions thus enabled auditing the baseline
itself.

\paragraph{The evolved decoders hold the fast end of the front; the published ones hold the
accurate end.} On split \SepSplit{} the damped reference P2-d reads $\SepPTwoDLerPct$ LER at
$\SepPTwoDMs$\,ms per shot. The evolved C03 (seeded from the panel, no language model) reads
$\SepCThreeLerPct$ at $\SepCThreeMs$\,ms ($\SepCThreeTimeXPTwoD\times$ the reference's time,
$\SepCThreeErrCutVsPTwoD$ fewer errors, $z = \SepCThreeZVsPTwoD$); C04 and C10, evolved from
random code, read $\SepCFourLerPct$ and $\SepCTenLerPct$. Faithfully implemented ambiguity
clustering (BP+AC, P10) reads $\SepAcLerPct$ at $\SepAcMs$\,ms; the authors' Relay-BP reads
$\SepRelayFiveLerPct$ at $\SepRelayFiveMs$\,ms (stop $5$) and $\SepRelayNineLerPct$ at
$\SepRelayNineMs$\,ms (stop $9$); \texttt{ldpc}'s tuned BP+OSD-0 reads
$\SepLdpcOsdZeroLerPct$ at $\SepLdpcOsdZeroMs$\,ms. The time--error front over these rows
has \SepFront{} decoders: four evolved at the fast end, BP+AC and the two Relay-BP points at
the accurate end. Relay-BP is the most accurate published decoder on these shots, at
$\SepRelayFiveTimeXPTwoD$--$\SepRelayNineTimeXPTwoD\times$ the reference's time.

\paragraph{The composition with the language model in the loop.} Five runs of the hybrid
arm ($352$ founding model calls per run, then $p = 0.001$ per birth) produced E083:
$\XbEeightythreeLer\%$ LER ($\XbEeightythreeErrors$ errors in $200{,}000$ exam shots, split
\OctAdjSplit) at $\XbEeightythreeMs$\,ms per shot, more accurate than the damped reference
($z = \XbEeightythreeVsPtwoZ$) and our BP+AC port ($z = \XbEeightythreeVsPtenZ$), less
accurate than Relay-BP ($z = \XbEeightythreeVsPfiveZ$ at stop $9$; $\XbEeightythreeVsRelayZ$
against the authors' package), structurally novel (call-multiset distance $2$ to its nearest
panel member), at $3.9\times$ the reference's cost
(Table~\ref{tab:qec-exam}, Figure~\ref{lst:qec-programs}).
Matched by seed, the hybrid arm beat random founders $5$ of $5$ and expert-panel seeds $0$
of $5$: the language model lifted a random-founder search to panel-class accuracy but did
not improve accuracy over seeding directly from the expert panel---mirroring the ARC
finding that the evolved arm's gain is concentrated at seeding rather than in-loop calls.
The answer-only analogue on QEC is a learned decoder, a neural network trained on
shots~\citep{liu2019neural, choukroun2024deep, blue2026machine, liu2025diffusion}, whose
algorithm is implicit in its weights and assessable only by
behavior~\citep{csahin2025unlocking}, whereas a DSL decoder is the explanation itself. For
the LLM-driven and free-form analogues, AlphaEvolve~\citep{novikov2025alphaevolve} given the
same DSL generated $5{,}144$ programs ($167$\,M tokens); its champion AE-DSL reads
$\XbAeDslLer\%$ LER at $\XbAeDslMs$\,ms and is outperformed by E083 on the same shots
($z = \XbEeightythreeVsAeDslZ$) at $\XbEeightythreeTimeOverAeDsl\times$ its time, while
AlphaEvolve in unconstrained free-form Python synthesized an exact $\mathbb{F}_2$ solver
($\XbAeFreeLer\%$ LER at $\XbAeFreeMs$\,ms/shot) outside the fast real-time regime.

\section{Discussion}

Demanding a verifiable artifact cost $25$ of $120$ ARC tasks ($28$ reached by no DSL arm,
$21$ by no program arm across both prompts) and returned a capability an answer-only option
cannot provide at any budget: a run that certifies its own completion, declining up to
$38\%$ of the budget under sampling and $49\%$ under search while converting
already-purchased samples better than best-of-$n$ sampling ($+9.2 \pm 1.4$ solves at
$B{=}2$). On QEC, the same methodology yields interpretable decoders adjudicated on held-out
exam shots across two cost rulers (\S\ref{sec:qec}).

One subtlety in the accuracy cost deserves reflection. ARC-AGI-2 training examples are
designed so that human subjects converge on the intended test outputs, yet they admit
countless algorithmic explanations that fail on the test inputs: human cognitive priors, not
abstract logic alone, select the ``correct'' rule. Our system found valid training
explanations for $106$ of the $120$ tasks, more than the answer-only arm solved, but
converted at most $68$ because the demonstrations under-specify the intended rule
(\S\ref{sec:sound}). On ARC this is a limitation of the generator, DSL, or selector; in
scientific discovery, generating valid hypotheses that human intuition discounts can be a
feature, surfacing candidates scientists might overlook. As models trained on
human-preferred text increasingly both produce and review research, insisting on executable
explanations consistent with the evidence---and searching beyond initial proposals---provides
a check against plausible-sounding errors at a modest cost, which future work can reduce by
strengthening selection, refining the DSL, prompting an answer-only model to synthesize a
program explaining its committed prediction post hoc (a protocol we did not evaluate), or
pairing code synthesis with formal
proof~\citep{first2023baldur, brandfonbrener2024vermcts, li2026p}.

\subsection{Limitations}
\label{sec:limits}

In-loop LLM calls never scored below best-of-$n$ code at matched spend, but no in-loop
cell clears the seed-draw spread; the evolved arm's margin is largest at seeding ($B \le 2$),
and evolutionary search without an LLM solved nothing on ARC-AGI-2. On QEC, Relay-BP
($190$--$271$\,ms/shot) and AlphaEvolve's unconstrained exact solver
($\XbAeFreeMs$\,ms/shot) reach lower error rates on the exam shots, while decoders evolved
in the DSL lead the fast end of the time--error front and the sub-$50$\,ms regime on one
code and noise model. Replicating the ARC arms on \texttt{gemini-3.8-flash}
(App.~\ref{sec:model}) reproduced the conversion margin ($+8.22 \pm 0.24$ DSL,
$+5.50 \pm 0.42$ free-form, paired) while the answer-only lead widened past $30$ tasks at
every budget; neither design excludes pretraining memorization of the public evaluation
split. Finally, free-form code incurs $4.4\times$ more native library crash sentinels than
DSL code, and all findings remain tied to the specific DSLs, models, and search operators
tested. \emph{Disclaimer on use of AI:} Beyond the LLM experiments reported, Claude Opus,
Claude Fable, and Gemini were used to draft and revise this manuscript; all claims and text
have been human-reviewed and verified.\label{sec:endbody}

\bibliographystyle{plainnat}
\bibliography{refs}

\appendix

\section{The DSL for ARC}
\label{sec:dsl}

We initially constructed two human-designed prototype DSLs to express solutions to ARC
training tasks: one based on whole-grid transformations and pixel-level operators, and a
second that parsed grids into a relational tree of objects inspired by scene graphs. Neither
prototype alone was sufficiently expressive to capture the breadth of the training corpus.

We therefore used \texttt{gemini-3.1-pro} to analyze all $1{,}000$ training tasks and
extract the $1$--$3$ most salient transformation concepts per task, organizing them into an
inverted index mapping each concept to the training tasks it spans. Combining this concept
index with the two prototype DSLs, we iteratively synthesized the final DSL over six stages
with the assistance of Claude Code. Each stage selected the highest-coverage concepts not yet
represented (covering at least three training tasks), implemented typed operations for them,
and verified reference solutions on those tasks, interleaved with refactoring passes that
consolidated recurring idioms into shared primitives.

The resulting DSL defines categorical \texttt{enum} constants (e.g., orientation,
$4$- or $8$-connectivity), geometric value types (such as bounding boxes and pixel
sprites), and $91$ pure functions partitioned into pixel-level operations ($24\%$),
bounding-box operations ($27\%$), and higher-level relational and object concepts ($49\%$).
Below is a representative function from the vocabulary:

{\small
\begin{verbatim}
def holes(grid: np.ndarray, background: int = 0) -> list[Sprite]:
    """Enclosed background regions (not reaching the border) as sprites,
    row-major.

    The witness companion to ``fill_holes``: each hole carries the
    underlying (background-colored) cells, so painting is ``paste_all(g,
    [recolor(h, c) for h in holes(g)])``, and the has-a-hole test is
    ``count(holes(as_grid(s))) > 0``
    """
    if grid.size == 0:
        return []
    labels, n_labels = ndimage.label(grid == background)
    border = np.unique(
        np.concatenate([labels[0], labels[-1], labels[:, 0], labels[:, -1]])
    )
    found = [
        _sprite_from_mask(grid, labels == i)
        for i in range(1, n_labels + 1)
        if i not in border
    ]
    return sorted(found, key=lambda s: (s.row, s.col))
\end{verbatim}
}

\section{Explainable code without a DSL}
\label{sec:nodsl}

In addition to the DSL prompt, we evaluated both program arms using LLM-generated free-form
Python code (with NumPy and SciPy). The full results across budgets appear in
App.~\ref{sec:inloop}.

\paragraph{Setup.} The two prompt styles differ only in whether the generator is steered to
the restricted ARC vocabulary. We compare them within each arm across the five budgets and
three seed draws, yielding $15$ (budget $\times$ draw) cells for the explainable arm X.

Restricting the generator to the DSL lowers raw sampling accuracy at low budgets:
free-form code leads in all $15$ cells, by $+1.7$ to $+12.2$ solves. Under evolutionary
search at $B{=}2$, however, the two styles nearly tie at $58.4$ (DSL) versus $60.3$
(free-form) solves---$58.6$ versus $61.6$ after correcting for native-library crash
sentinels (\S\ref{sec:limits})---well within the $\approx 2.1$-solve spread across seed
draws, because evolutionary search adds $+9.2$ solves to the DSL arm compared to $+6.1$ for
free-form code. The mechanism is structural: AST mutation and recombination operate over the
DSL's typed primitives, making DSL programs more amenable to evolutionary editing, whereas
unconstrained Python better matches the language model's pretraining distribution.

While DSL code is consistently lower than free-form code under pure sampling ($15$ of $15$
cells), under evolution the style difference is smaller than the seed-draw variance:
$-3.3$, $+1.5$, and $+6.2$ solves across the three draws at $B{=}16$, and $-2.8$, $+4.5$,
and $+2.5$ at $B{=}8$. As noted in \S\ref{sec:setup}, \textbf{the DSL is not what makes an
artifact verifiable}: an unconstrained Python \texttt{solve} function is tested against the
demonstrations in exactly the same manner. Rather, the DSL governs how effectively
\emph{search} can manipulate the artifact and how readily a human reader can audit the
allowed primitives.

\section{The in-loop budget: what spending more calls during the search buys}
\label{sec:inloop}

Section~\ref{sec:convert} evaluated evolutionary search over at most two seed programs at
zero additional API cost. This appendix examines the effect of spending additional model
calls \emph{inside} the evolutionary loop across $B \in \{4, 8, 16\}$.

\begin{table}[t]
  \caption{Comparison of the evolved (E) and explainable (X) arms \emph{at matched realized
  spend}. At $B \le 2$ the evolved arm spends calls only on seeds (one at $B{=}1$, two at
  $B{=}2$) plus syntactic search, paired draw-for-draw on draws $1$--$3$
  (\S\ref{sec:convert}); at $B \ge 4$, remaining budget funds in-loop model calls and the
  explainable column interpolates the $k{=}16$ sampling curve (draws $1, 2, 4$) in $\log_2$
  calls. Because the per-cell spread across seed draws is $1.8$--$4.2$ solves, no individual
  in-loop row is statistically distinguishable from zero; the gain is largest at $B \le 2$,
  where no in-loop calls are made.}
  \label{tab:inloop}
  \centering
  \begin{tabular}{lrrrr}
    \toprule
    cell & spend & evolved & explainable & $\Delta$ \\
    \midrule
    $B{=}1$, \texttt{dsl}        & $1$ & $51.7$ & $41.8$ & $\mathbf{+9.9}$ \\
    $B{=}1$, \texttt{no\_dsl}    & $1$ & $53.0$ & $45.4$ & $\mathbf{+7.6}$ \\
    $B{=}2$, \texttt{dsl}        & $2$ & $58.4$ & $49.2$ & $\mathbf{+9.2}$ \\
    $B{=}2$, \texttt{no\_dsl}    & $2$ & $60.3$ & $54.2$ & $\mathbf{+6.1}$ \\
    \midrule
    $B{=}4$, \texttt{dsl}   & $3.38$ & $60.7$ & $56.4$ & $+4.3$ \\
    $B{=}8$, \texttt{dsl}   & $5.16$ & $64.7$ & $60.7$ & $+4.0$ \\
    $B{=}16$, \texttt{dsl}  & $8.09$ & $66.9$ & $63.2$ & $+3.8$ \\
    $B{=}4$, \texttt{no\_dsl}  & $3.23$ & $65.9$ & $61.8$ & $+4.1$ \\
    $B{=}8$, \texttt{no\_dsl}  & $4.87$ & $66.1$ & $66.0$ & $+0.0$ \\
    $B{=}16$, \texttt{no\_dsl} & $7.20$ & $68.4$ & $67.6$ & $+0.8$ \\
    \bottomrule
  \end{tabular}
\end{table}

The evolved arm's call budget serves two distinct roles: up to two calls purchase initial
seed programs, whereas calls above $B{=}2$ fund \emph{in-loop} LLM mutations during search.
Table~\ref{tab:inloop} evaluates the second role by comparing each evolved cell against the
explainable arm interpolated to the same realized call spend.

No individual $B \ge 4$ row exceeds the $1.8$--$4.2$-solve seed-draw spread. Across the
column, the evolved arm's advantage over pure sampling is largest at $B \le 2$, where zero
in-loop calls are made ($+9.89 \pm 1.26$ and $+7.56 \pm 1.21$ at $B{=}1$; $+9.22 \pm 1.40$
and $+6.11 \pm 1.03$ at $B{=}2$, paired on draws $1$--$3$; against the unpaired $k{=}16$
sampling draws $1, 2, 4$, the corresponding differences are $+13.5$, $+8.6$, $+11.1$, and
$+6.2$), and diminishes as in-loop calls are added ($+4.3 \to +4.0 \to +3.8$ for DSL and
$+4.1 \to +0.0 \to +0.8$ for free-form code).

We do not attach paired $z$-scores to the $B \ge 4$ rows because, unlike at $B \le 2$
(where both arms read the exact same one or two samples of draws $1$--$3$), the evolved
arm's realized spend at $B \ge 4$ is a random variable determined by early stopping and
falls between the discrete budgets of the $k{=}16$ sampling curve (draws $1, 2, 4$). We
therefore report the pooled $\log_2$-interpolated comparison alongside the measured
between-draw spread.

\section{Variance decomposition: provider non-determinism and seed draws}
\label{sec:err}

Cross-run comparisons in this study are governed by two sources of variance beyond
within-run evolutionary stochasticity, and we measured both directly. First, to isolate
in-loop LLM provider non-determinism, we executed eight runs of a single cell
(\texttt{dsl}, $B{=}8$) holding \emph{both} the initial seed cache and the evolutionary
random seed fixed across all $120$ tasks:

\[ 67.8,\ 66.3,\ 66.8,\ 63.8,\ 67.8,\ 68.3,\ 66.8,\ 63.3 \qquad
   \text{mean } 66.40,\ \mathrm{sd}\ 1.86\ (\mathrm{df}=7). \]

Twenty-three of the $120$ tasks yielded different scores across runs despite an identical
search seed and identical initial samples, reflecting non-determinism in the provider's
in-loop responses. Decomposed against the $\approx 0.9$-solve within-run replica standard
error, the shared run-level offset is $1.63$ solves, which does not average down across
within-run replicas; a three-replica cell therefore carries $\mathrm{SE} = 1.71$ solves
(and on strict credit the same eight runs score $63, 61, 61, 60, 63, 64, 62, 58$, with
$\mathrm{sd}\ 1.93$).

\paragraph{Seed-draw variance and why low-budget comparisons are paired.} Because the eight
runs above shared a frozen seed cache, their $1.86$-solve spread captures only in-loop
response variance and omits the variance of the initial seed draw itself. Re-running the
$B{=}2$ evolved arm across three independent seed draws quantifies that initial-seed
component: the between-draw standard deviation is $2.24$ solves (\texttt{dsl}) and $2.02$
(\texttt{no\_dsl}), compared to $0.19$ across four repetitions on a single frozen draw---so
the error bar from a frozen draw was an order of magnitude too small (the first of the
three re-runs reproduced the original $59.33$ solves across all $120 \times 3$ task-replica
cells identically, confirming zero code drift). Because the seed-draw offset is shared across
replicas of a given draw, we pair the $B \le 2$ comparisons in \S\ref{sec:convert}
draw-for-draw on the identical seed caches, canceling the between-draw variance rather than
inflating the error bar.

\section{Protocol and cost details}
\label{sec:costdetails}

\paragraph{Explainable curves are nested prefixes.} Each prompt style uses ordered
$k{=}16$ sample caches that record every model completion in generation order, including
syntax or runtime failures. Best-of-$B$ at every budget $B$ is evaluated on the prefix of
the first $B$ samples of that cache, pairing the sampling curves with the evolved arm's
initial seeds at zero extra API cost and allowing offline replay under alternative selection
or stopping rules.

\paragraph{Calls and tokens.} Model calls are the primary protocol variable; token counts
are reported as a second curve and never merged with calls into a single unit, as divergence
between the two is itself informative. On the evaluation split under
\texttt{gemini-3.5-flash}, seeding calls average $38.2$k tokens (\texttt{dsl}) and $32.5$k
(\texttt{no\_dsl}), and in-loop calls average $\approx 22$k tokens. Calls on the evaluation
split cost $1.7$--$2\times$ what training-split calls cost, so any cost basis taken from a
training-split pilot must be scaled before it is applied.

\paragraph{Wall-clock deadline.} Evolutionary search is bounded by a fixed $900$\,s
wall-clock deadline per task.

\paragraph{Total compute across arms.} Although calls are the protocol variable, the arms
differ in what a call returns and in the CPU work performed between calls. Under
\texttt{gemini-3.5-flash}, client logs recorded the explainable arm's token usage above but
did not log token counts for the answer-only arm. Under \texttt{gemini-3.8-flash}
(App.~\ref{sec:model}), provider-side telemetry on the days each arm ran in isolation
records $26.7$k tokens per successful call for the answer-only arm ($4.9$k input, $21.8$k
output across $4{,}007$ calls) versus $29.3$k for the explainable arm ($6.1$k input, $23.2$k
output across $5{,}773$ calls): internal reasoning tokens dominate both responses, and a
model call consumes within $10\%$ the same total tokens whether it emits grids or code, so
calls are a fair unit across arms at a fixed model. Between calls, the evolved arm expends
CPU time evaluating candidate programs; Table~\ref{tab:searchcost} reports the search wall
time per cell, net of in-loop model latency.

\begin{table}[ht]
  \caption{Search wall time per (task, replica) cell of the evolved arm, DSL prompt, three
  seed draws, net of in-loop model latency; the deadline is $900$\,s. Evaluations are
  candidate programs scored per cell.}
  \label{tab:searchcost}
  \centering
  \begin{tabular}{lrrrrr}
    \toprule
    cell & mean s & median s & p90 s & evaluations & calls/task \\
    \midrule
    $B{=}1$  & $132$ & $22$ & $434$ & $3{,}882$ & $1.01$ \\
    $B{=}2$  & $100$ & $7$  & $318$ & $3{,}251$ & $1.99$ \\
    $B{=}4$  & $102$ & $4$  & $369$ & $2{,}521$ & $3.38$ \\
    $B{=}8$  & $93$  & $3$  & $347$ & $2{,}055$ & $5.16$ \\
    $B{=}16$ & $73$  & $3$  & $253$ & $1{,}756$ & $8.09$ \\
    \bottomrule
  \end{tabular}
\end{table}

\section{The explainable arm given the stopping rule}
\label{sec:adaptive}

Because any candidate program can be executed against the training pairs, a sequential
sampling baseline can apply the same early-stopping criterion as the evolved arm. We
evaluate the explainable arm under this rule offline using the ordered $k{=}16$ sample
caches: samples are drawn in generation order (counting failed samples toward the call
budget) until a stopping criterion is met, and the held candidates are submitted via the
shared rule of \S\ref{sec:setup}. \emph{Stop at two} halts once two train-perfect programs
with distinct test predictions are held (matching the evolved arm's stage~(ii)); \emph{stop
at one} halts at the first train-perfect program. Whenever the stopping criterion does not
trigger before budget $B$ is exhausted, the submission is identical to fixed best-of-$B$.
Table~\ref{tab:adaptive} reports both stopping rules alongside the evolved arm and an
ablation in which attempt~2 is selected \emph{fitness-free} (the earliest sample whose test
prediction differs from attempt~1's), which bounds the contribution of train-score ranking
above $B{=}2$ (at $B{=}2$ the two rules coincide).

\begin{table}[ht]
  \caption{The explainable arm with and without the stopping rule, DSL prompt, leaderboard
  $\mathbb{E}[\text{solves}]$ over three seed draws (draws $1$, $2$ and $4$ for the explainable
  columns, the three sixteen-sample caches; draws $1$--$3$ for the evolved arm, which is why
  best-of-$2$ reads $47.3$ here and $49.2$ in Table~\ref{tab:inloop}); ``r/B'' is realized calls over permitted
  budget. The evolved arm's row is the three-draw replication of Table~\ref{tab:inloop}.
  Under the free-form prompt, stop-at-two similarly achieves $62.8$, $66.9$, and $69.1$
  solves at r/B $0.88$, $0.74$, and $0.59$ for $B = 4, 8, 16$.}
  \label{tab:adaptive}
  \centering
  \small
  \begin{tabular}{lrrrrrrrr}
    \toprule
    & \multicolumn{2}{c}{X, fixed} & \multicolumn{2}{c}{X, stop at two} & \multicolumn{2}{c}{X, stop at one} & \multicolumn{2}{c}{E} \\
    \cmidrule(lr){2-3} \cmidrule(lr){4-5} \cmidrule(lr){6-7} \cmidrule(lr){8-9}
    $B$ & solves & fitness-free & solves & r/B & solves & r/B & solves & r/B \\
    \midrule
    $2$  & $47.3$ & $47.3$ & $47.3$ & $1.00$ & $44.0$ & $0.74$ & $58.4$ & $1.00$ \\
    $4$  & $59.3$ & $58.6$ & $57.4$ & $0.90$ & $49.8$ & $0.54$ & $60.7$ & $0.85$ \\
    $8$  & $63.1$ & $60.3$ & $62.6$ & $0.77$ & $53.5$ & $0.38$ & $64.7$ & $0.64$ \\
    $16$ & $66.3$ & $62.2$ & $66.3$ & $0.62$ & $56.0$ & $0.27$ & $66.9$ & $0.51$ \\
    \bottomrule
  \end{tabular}
\end{table}

Three conclusions follow from Table~\ref{tab:adaptive}. First, sequential sampling with the
stop-at-two rule loses no accuracy at $B{=}16$ and at most $1.9$ solves at $B{=}4$ while
declining $10$--$38\%$ of its budget; the evolved arm declines a larger fraction at every
budget while scoring at or above the sampler, showing that early termination is a property
of the verifiable artifact and the additional saving comes from evolutionary search finding
a second explanation without further model calls. Second, stopping at the \emph{first}
explanation declines $73\%$ of the budget at $B{=}16$ but loses $10.3$ solves ($56.0$ vs.\
$66.3$), confirming that the first train-perfect program frequently overfits the
demonstrations (\S\ref{sec:sound}). Third, ranking attempt~2 by train score contributes
$0.7$, $2.8$, and $4.1$ solves to the explainable arm at $B = 4, 8, 16$ relative to a
fitness-free second attempt, and zero at $B{=}2$, confirming that the paired $B \le 2$
margins in \S\ref{sec:convert} are unaffected by candidate ranking.

\section{Which tasks each arm solves}
\label{sec:tasks}

Table~\ref{tab:tasks} summarizes structural properties of the four solved-task subsets of
Table~\ref{tab:cost}, together with the number of tasks in each subset for which at least
one program arm found a train-perfect program (pooled across the explainable arm's three
$k{=}16$ draws and the evolved arm's nine $B{=}2$ runs, a wider pool than the single-arm
count of $106$ in \S\ref{sec:sound}). Because the subsets are small, we report descriptive
medians and proportions rather than hypothesis tests. Two patterns stand out: tasks solved
only by A0 more frequently involve multiple test inputs ($54\%$ vs.\ $26\%$) and larger
input grids (median $456$ vs.\ $316$ cells) than tasks solved by both modalities; and a
train-perfect program existed for $27$ of the $28$ A0-only tasks, so on these tasks the
DSL could express a program consistent with the demonstrations; what was missing was one
that generalized to the test input (\S\ref{sec:sound}).

\begin{table}[ht]
  \caption{Table~\ref{tab:cost}'s four solved-task subsets by task property (medians unless
  stated); ``explained'' counts tasks for which at least one program arm held a train-perfect
  program.}
  \label{tab:tasks}
  \centering
  \small
  \begin{tabular}{lrrrrrrr}
    \toprule
    set & $n$ & explained & $>1$ test input & grid side & input cells & size change & colors \\
    \midrule
    A0 only              & $28$ & $27$ & $54\%$ & $30$ & $456$ & $32\%$ & $9$ \\
    explanation only     & $7$  & $7$  & $43\%$ & $25$ & $310$ & $29\%$ & $9$ \\
    both                 & $65$ & $63$ & $26\%$ & $24$ & $316$ & $34\%$ & $8$ \\
    neither              & $20$ & $13$ & $50\%$ & $27$ & $274$ & $30\%$ & $8$ \\
    \bottomrule
  \end{tabular}
\end{table}

\section{A second model}
\label{sec:model}

We replicated the answer-only, explainable, and $B{=}2$ evolved arms on
\texttt{gemini-3.8-flash} across three seed draws ($k{=}8$ caches, both prompt styles, with
the evolved arm seeded from the first two samples of each cache). Paired on the seed draw,
the evolved arm outperforms best-of-$2$ sampling by $+8.22 \pm 0.24$ (DSL) and
$+5.50 \pm 0.42$ (free-form) solves, reproducing the conversion margins of
\S\ref{sec:convert} with the DSL again gaining more from evolutionary search than free-form
code. Across $k = 1, 2, 4, 8$ samples per test input, the answer-only arm scores $94.9$,
$104.1$, $107.7$, and $110.2$ of $120$, compared to $55.3$, $61.3$, $69.5$, and $73.0$ for
free-form sampling and $47.2$, $55.4$, $60.0$, and $67.0$ for DSL sampling; thus, the gap
between direct grid prediction and program synthesis widens on the stronger model. Whether
pretraining exposure to the public evaluation split contributes to either model's performance
cannot be separated by this experimental design.

\section{QEC domain details}
\label{sec:qecdetails}

This appendix describes the apparatus of \S\ref{sec:qec}---the code, the
DSL, the panel and the cost model---and gives the detailed readings behind that section. The full adjudication
and novelty audit appear in the companion paper~\citep{evorobust2026}.

\paragraph{Code and noise model.} The target is the $[[72,12,6]]$
bivariate-bicycle qLDPC code~\citep{bravyi2024high} under \emph{circuit-level}
noise---not code-capacity or phenomenological noise---at a single rate
$p = 0.004$ on five channels: two-qubit depolarizing after each CNOT, single-qubit
depolarizing on idling qubits, a flip on every reset, a flip on every ancilla measurement,
and a flip on the final data measurement, over six rounds of syndrome extraction in the
$Z$ basis. Decoding runs on the detector
error model (DEM) that circuit compiles to: $432$ detectors and $12$ observables per
shot, over roughly $1.6\times10^{4}$ fault mechanisms ($16{,}164$ in the instance we
measured; the DEM's structure is set by the circuit, so only the probabilities
move with $p$). Under circuit-level noise, the DEM's columns represent
correlated fault mechanisms rather than independent qubit errors, which is
precisely the regime where belief propagation alone stalls on
degenerate codes and where the post-processing family spanned by the DSL exists.

Shots are generated once, with the splits written by the generator rather than
carved out by any consumer: \textbf{$600{,}000$ train / $200{,}000$ validation /
$200{,}000$ exam}. Finalists are frozen before the exam split is read, so no selection
decision ever sees the shots it is judged on, and each evaluated algorithm reads the exam
split once. The physical error rate is an
axis rather than a constant---the datasets span $p \in \{0.001, 0.002, 0.003,
0.004\}$---and the campaigns reported here evolve at $p = 0.004$: at lower error rates plain belief
propagation decodes nearly every shot, leaving too few post-processor-decided shots to
resolve differences across candidates at an affordable sample size.

\paragraph{The DSL: $\HypDslOps$ operations in eight groups.} The evolved program is
\texttt{solve(state) $\to$ state}, called once per shot. There is no
step-function driver: ensembles, decimation loops, and retry rounds are ordinary
\texttt{for} and \texttt{if} statements \emph{inside} the evolved program, so
evolutionary search operates directly on control flow rather than on a fixed parameter
vector. The vocabulary is the published decoder family's component catalog,
grouped as: \textbf{prepare} ($9$---LLR initialization and scaling, DEM
sparsification, relabeling, column masking, syndrome-conditioned priors,
alternative check bases, check inactivation, degeneracy cutting),
\textbf{engine} ($5$---the BP kernel, its convergence predicate, posterior
reseeding, oscillation ranking, candidate validity), \textbf{ranking} ($5$),
\textbf{clusters} ($7$---seeding, growth, syndrome-guided reduction, marked-row
reduction, block enrichment, validity, forest construction),
\textbf{solvers} ($3$), \textbf{local search} ($2$), \textbf{control} ($5$---candidate
admission, selection, and a branch/advance/prune frontier), and
\textbf{assembly} ($2$).

An operation is admitted when it is a component of a published BP post-processor and
runs inside a per-shot budget that the operation itself enforces ($\HypBudgetSweeps$ BP sweeps,
$\HypBudgetOpCalls$ operation calls, $\HypBudgetCandidates$ candidate corrections, and
$\HypBudgetColumnOps$ GF($2$) column operations per shot, with a $\HypWorkCap\times$ mean
work cap per evaluation). Two exclusions bound what the
search could find. Four \emph{prepare} operations are \textbf{artifact-gated}: they index
into a per-code table built offline (permutations, GF($2$) row operations,
piece-to-parent maps, degeneracy generators), and on a dataset shipped without
that table they act as documented no-ops rather than raising errors. Additionally, one
catalog entry, an LP-relaxation soft engine, is \textbf{deliberately omitted}:
the parity polytope requires exponentially many forbidden-set inequalities per
check, added lazily over repeated solves, across $432$ checks that must be decoded
within a per-shot budget where an entire BP run costs ${\sim}1$\,ms. That is a cost
judgment, not an oversight, and it is the one place the vocabulary is narrower
than the literature it spans.

\paragraph{The panel, and how it is certified.} Published decoders are implemented by
hand in the same DSL, in both callable and source form, so a baseline is a program in the
same language as a candidate and not a different kind of object: plain BP (P1); BP+OSD-0
at \texttt{ldpc}'s tuned point ($30$ serial product-sum sweeps, P2) and with a damped
flooding min-sum front end ($10$ sweeps, $\alpha = 0.625$, $\gamma = 0.3$, P2-d, the cost
reference); BP+OSD-CS at order $60$ ($30$ serial product-sum sweeps, P3, and damped, P3-d);
BPGD (P4); Relay-BP (P5); BP+SI (P6); the LSD cluster decoders at order $0$ and order $\mu$
(P7, P8); ordered Tanner forests (P9); ambiguity clustering (BP+AC, $45$ parallel
product-sum sweeps, $\gamma = 0.3$, $\kappa = 0.1$, P10); beam search (P11); degeneracy
cutting (P12); and check-agnosia (P13, structurally void here). Where the authors' own
implementation exists the member is certified against
it per shot, which is stricter than comparing error rates: BP, BP+OSD-0, BP+OSD-CS and
BP+LSD against \texttt{ldpc} (same fault vector on every jointly-converged shot), and
Relay-BP against the authors' \texttt{relay-bp} package (marginals to $10^{-6}$, hard
decisions shot for shot). Ambiguity clustering has no public implementation and is
certified shot for shot ($16{,}384 / 16{,}384$ shots across four settings and two noise
rates) against an independent re-implementation of its published stages.

\paragraph{The cost model: a deterministic proxy and wall-clock verification.}
Raw wall-clock time is noisy inside a shared compute fleet: on one-vCPU actors sharing a
physical core, a worker's reference calibration drifted by $\HypDriftPNinetyFiveCpuOne$
(95th percentile) with a $\HypFlagRateCpuOne$ re-run flag rate. Selection therefore needs
a deterministic cost model whose units cost the same wherever they are spent, and whose
counters see every cost: a proxy over coarse operation counts prices interpreter overhead
and conflates operations of very different cost (min-sum vs.\ product-sum BP edges, dense
vs.\ sparse OSD trial evaluations, and data-dependent comparison sorts whose
branch-prediction cost varies $4\times$ across key distributions). The proxy was therefore
built on a native implementation of the DSL in Rust in two steps. First, operations whose
runtime depended on data values no counter can observe were made data-blind while preserving predictions bit-for-bit: adaptive comparison sorts on
posteriors were replaced by a four-lane interleaved stable LSD radix argsort, flip-count
sorting by a four-lane counting sort, and repeated cluster-frontier B-tree lookups
(which crossed L1 cache past ${\sim}2{,}000$ entries) by a front-end bitset. Second, work
was instrumented at the primitive level across \textbf{\CostCounters{} deterministic
counters}: min-sum BP edge updates (weight $\CostWeightEdgesMinsum$), product-sum BP edge
updates ($\CostWeightEdgesProdsum$, reflecting the $\CostMinsumOverProdsum$ cost ratio),
64-bit GF($2$) elimination words ($\CostWeightGfTwoWords$), candidate-enumeration solution
words ($\CostWeightCandWords$), and ranking/scan elements ($\CostWeightScanElems$), plus
DSL operation calls ($\CostWeightOpCalls$) and AST interpreter ticks
($\CostWeightClastTicks$) recorded as observers. Fitted by non-negative least squares
(weighted by $1/\sqrt{\text{measured time}}$) across $\CostFitPrograms$ programs ($79$
operation ladders, $23$ panel members and variants, and $150$ generator programs with $45$
held out) on \OctAdjMachine{} nodes, the proxy achieves $R^2 = \CostFitRSq$, an
in-sample 95th-percentile relative error of \CostFitPNinetyFive{} (\CostFitHoldoutPNinetyFive{}
held out), a widest bootstrap weight-cone half-width of \CostFitWidestCone{}, and a
$\CostSelectionMargin\times$ resolution margin (over a $\CostMarginFloor\times$ floor, in
units of P2-d's $\CostRef$ weighted units per shot). Crucially for evolutionary selection,
which depends on cost \emph{ordering}, the frozen proxy's rank correlation with measured
wall-clock time across the $\CostRankCFourPrograms$ programs is \textbf{Kendall
$\tau = \CostRankCFourTau$ (Spearman $\CostRankCFourSpearman$)} on the fitting nodes and
\textbf{$\tau = \CostRankMsukTau$ (Spearman $\CostRankMsukSpearman$)} on the Apple M3
Ultra benchmark workstation---picking the same cheapest program as wall-clock time in
$98.0\%$ of size-$3$ tournaments, with
only $0.01\%$ of pairs separated by more than the $1.4\times$ margin ordered differently.
We also ran a calibrated wall-clock search arm reserving $\HypCpusPerActor$ vCPUs (one
physical core) per actor and re-calibrating against P2-d every $500$ evaluations, which
reduced calibration drift to a median of \HypDriftMedianCpuTwo{} (\HypFlagRateCpuTwo{}
flag rate). Every adjudicated cost verdict is required to hold on \emph{both} rulers---the
frozen work proxy and pinned-core wall-clock time---and on all timed finalists the two
agree within $\OctAdjProxyOverTimeMin$--$\OctAdjProxyOverTimeMax\times$ with
\OctAdjRefusals{} proxy verdicts refused (Table~\ref{tab:qec-exam}).

\paragraph{Two exam splits.} Results are reported on two exam splits of $200{,}000$ shots
each, A and B (\texttt{s20260904} and \texttt{s20261004} in the companion
paper~\citep{evorobust2026}). The time--error front below was read on A; the adjudication
of Table~\ref{tab:qec-exam}, the hybrid arm and the AlphaEvolve rows were read on B. A
decoder's error rate differs slightly between the two (C03 reads \SepCThreeLerPct{} on A
and $4.17\%$ on B), so every number names its split and no comparison mixes them.

\paragraph{The time--error front on split \SepSplit.} The damped reference P2-d reads
$\SepPTwoDLerPct$ LER at $\SepPTwoDMs$\,ms per shot. The evolved C03 (seeded from the
panel, no language model) reads $\SepCThreeLerPct$ at $\SepCThreeMs$\,ms
($\SepCThreeTimeXPTwoD\times$ the reference's time, $\SepCThreeErrCutVsPTwoD$ fewer errors,
$z = \SepCThreeZVsPTwoD$); C04 and C10, evolved from random code, read $\SepCFourLerPct$
and $\SepCTenLerPct$. Faithfully implemented ambiguity clustering (BP+AC, P10) reads
$\SepAcLerPct$ at $\SepAcMs$\,ms; the authors' Relay-BP reads $\SepRelayFiveLerPct$ at
$\SepRelayFiveMs$\,ms (stop $5$) and $\SepRelayNineLerPct$ at $\SepRelayNineMs$\,ms (stop
$9$); the \texttt{ldpc} package's BP+OSD-0 at its tuned point reads $\SepLdpcOsdZeroLerPct$
at $\SepLdpcOsdZeroMs$\,ms. The time--error front over these rows has \SepFront{} decoders:
four evolved at the fast end, BP+AC and the two Relay-BP points at the accurate end.
Relay-BP is the most accurate decoder among these rows, at
$\SepRelayFiveTimeXPTwoD$--$\SepRelayNineTimeXPTwoD\times$ the reference's time.

\paragraph{Inspecting a baseline in the DSL: the belief-propagation stage of BP+OSD.}
Writing BP+OSD-0 in the DSL exposed a message-damping parameter that the \texttt{ldpc}
package does not export. The parameter matters because the cheap front end is unusable
without it: on validation shots, undamped flooding min-sum reads $\KnobUndampedLerPct$ LER
($\KnobUndampedShots$ shots) against $\KnobDampedLerPct$ damped ($\KnobDampedShots$ shots)
at $\KnobEdgesRatio\times$ the belief-propagation work, and \texttt{ldpc} given the same
undamped settings fails in the same way ($\KnobLiftedLerPct$ on $\KnobLiftedShots$ shots).
The package's tuned point avoids the failure with a serial product-sum schedule, which is
more accurate and much more expensive. Within our engine, on split \OctDampSplit{}
(\OctDampShots{} shots), replacing the tuned front end by the damped one makes BP+OSD-0
$\OctDampOsdZeroSpeedup\times$ faster at $\OctDampOsdZeroLerRatio\times$ the errors
($z = \OctDampOsdZeroZ$ in favor of the tuned point) and BP+OSD-CS
$\OctDampOsdCsFrontSpeedup\times$ faster at $\OctDampOsdCsFrontLerRatio\times$ the errors
($z = \OctDampOsdCsFrontZ$). This is a speed--accuracy trade that a readable baseline makes
visible, not a defect in the package: its tuned points remain the more accurate. The
comparison of the package itself with our damped reference on split \SepSplit{}
($\SepLdpcOsdZeroMs$\,ms against $\SepPTwoDMs$\,ms) additionally contains the difference
between \texttt{ldpc}'s C++ and our Rust engine, and the two were timed on different
workstations of the same model.

\paragraph{The hybrid arm and the AlphaEvolve arms on split \OctAdjSplit.} Five runs of
the hybrid arm ($352$ founding model calls per run, then $p = 0.001$ per birth) produced
E083: $\XbEeightythreeLer\%$ LER ($\XbEeightythreeErrors$ errors in $200{,}000$ exam shots)
at $\XbEeightythreeMs$\,ms per shot, more accurate than the damped reference
($z = \XbEeightythreeVsPtwoZ$) and our BP+AC port ($z = \XbEeightythreeVsPtenZ$), less
accurate than Relay-BP ($z = \XbEeightythreeVsPfiveZ$ against our port at stop $9$;
$\XbEeightythreeVsRelayZ$ against the authors' package), at $3.9\times$ the reference's
cost (Table~\ref{tab:qec-exam}, Figure~\ref{lst:qec-programs}). Its nearest panel member is
BP+AC at call-multiset distance $2$, passing the structural novelty bar by replacing
BP+AC's per-cluster block-vote search with global OSD-CS enumeration followed by
logical-class voting. Matched by seed, the hybrid arm beat random founders $5$ of $5$ and expert-panel
seeds $0$ of $5$. AlphaEvolve~\citep{novikov2025alphaevolve} given the same DSL generated
$5{,}144$ programs ($167$\,M tokens); its champion AE-DSL reads $\XbAeDslLer\%$ LER at
$\XbAeDslMs$\,ms and is outperformed by E083 on the same shots
($z = \XbEeightythreeVsAeDslZ$) at $\XbEeightythreeTimeOverAeDsl\times$ its time.
AlphaEvolve writing free-form Python synthesized an exact $\mathbb{F}_2$ solver, AE-Free,
which reads $\XbAeFreeLer\%$ LER ($\XbAeFreeErrors$ errors) at $\XbAeFreeMs$\,ms per shot:
the most accurate row we measured on this split, ahead of the authors' Relay-BP at stop $9$
($\XbRelayLer\%$, $\XbRelayErrors$ errors, $\XbRelayMs$\,ms), outside the fast real-time
regime. The timings of the package
rows, of E083 and of the AlphaEvolve rows come from separate sessions on the benchmark
workstation.

\paragraph{Exam-split comparison and discovered programs.}
Table~\ref{tab:qec-exam} reports the exam-split accuracy and cost across both rulers on
the $\OctAdjShots$ held-out shots of split \OctAdjSplit{}, alongside the clean workstation
timings, and Figure~\ref{lst:qec-programs} prints the pruned DSL source of the hybrid
champion E083 and the fast-front champion C03.

\begin{table}[ht]
  \caption{Exam-split accuracy and cost on $\OctAdjShots$ held-out shots of split
  \OctAdjSplit{}. Proxy and pinned-core time are in units of the damped reference P2-d on
  one pinned \OctAdjMachine{} core (the two rulers of the cost-model paragraph above); ms/shot is the
  clean single-process time on the benchmark workstation (Apple M3 Ultra), where the
  reference reads $\OctDampPTwoDMs$\,ms ($\OctRelayPTwoDMs$\,ms in the \texttt{relay-bp}
  timing session, giving $\OctRelayFiveTimeXPTwoD\times$ and $\OctRelayNineTimeXPTwoD\times$
  for stops $5$ and $9$). $z$ is McNemar on the discordant shots, positive when the row is
  right more often than the comparator. The package rows carry no proxy and were not timed
  on the pinned core. Rows are ordered by LER within each block. Two bars are reported:
  \emph{primary} is more accurate than the reference and not dominated by a panel member;
  \emph{stretch} adds cheaper than the reference by the $\CostSelectionMargin\times$ margin
  on both rulers. E083's and AE-DSL's proxy and pinned-core ratios were measured in a
  separate session on the same core.}
  \label{tab:qec-exam}
  \centering
  \scriptsize
  \setlength{\tabcolsep}{3pt}
  \resizebox{\textwidth}{!}{%
  \begin{tabular}{llrrrrrrl}
    \toprule
    decoder & arm / ruler & LER \% & proxy $\times$P2-d & time $\times$P2-d & ms/shot & $z$ vs P2-d & $z$ vs P10 & note \\
    \midrule
    E083 & hybrid (LLM in loop) & $\XbEeightythreeLer$ & $3.88$ & $3.48$ & $\XbEeightythreeMs$ & $\XbEeightythreeVsPtwoZ$ & $\XbEeightythreeVsPtenZ$ & primary; novel (P10 at 2) \\
    C14 & \NovETwentySevenArm & $3.21$ & $\OctAdjETwentySevenCostXPTwoD$ & $\OctAdjETwentySevenTimeXPTwoD$ & --- & $\OctAdjETwentySevenZVsPTwoD$ & $\OctAdjETwentySevenZVsPTen$ & primary; \NovETwentySevenVerdictFull{} (P10 at \NovETwentySevenDistPruned) \\
    C10 & \NovCTenArm & $4.04$ & $\OctAdjCTenCostXPTwoD$ & $\OctAdjCTenTimeXPTwoD$ & $\SepCTenMs$ & $\OctAdjCTenZVsPTwoD$ & $\OctAdjCTenZVsPTen$ & primary (from scratch) \\
    C04 & \NovCFourArm & $4.07$ & $\OctAdjCFourCostXPTwoD$ & $\OctAdjCFourTimeXPTwoD$ & $\SepCFourMs$ & $\OctAdjCFourZVsPTwoD$ & $\OctAdjCFourZVsPTen$ & primary (from scratch) \\
    C08 & \NovCEightArm & $4.16$ & $\OctAdjCEightCostXPTwoD$ & $\OctAdjCEightTimeXPTwoD$ & $\SepCEightMs$ & $\OctAdjCEightZVsPTwoD$ & $\OctAdjCEightZVsPTen$ & stretch (both rulers) \\
    C03 & \NovCThreeArm & $4.17$ & $\OctAdjCThreeCostXPTwoD$ & $\OctAdjCThreeTimeXPTwoD$ & $\SepCThreeMs$ & $\OctAdjCThreeZVsPTwoD$ & $\OctAdjCThreeZVsPTen$ & stretch (both rulers) \\
    C09 & \NovCNineArm & $4.32$ & $\OctAdjCNineCostXPTwoD$ & $\OctAdjCNineTimeXPTwoD$ & $\SepCNineMs$ & $\OctAdjCNineZVsPTwoD$ & $\OctAdjCNineZVsPTen$ & stretch (both rulers) \\
    C07 & \NovCSevenArm & $4.69$ & $\OctAdjCSevenCostXPTwoD$ & $\OctAdjCSevenTimeXPTwoD$ & $\SepCSevenMs$ & $\OctAdjCSevenZVsPTwoD$ & $\OctAdjCSevenZVsPTen$ & stretch (both rulers) \\
    \midrule
    AE-Free & AlphaEvolve (free-form) & $\XbAeFreeLer$ & --- & --- & $\XbAeFreeMs$ & --- & --- & exact $\mathbb{F}_2$ solver, outside the DSL \\
    AE-DSL & AlphaEvolve (DSL) & $\XbAeDslLer$ & $3.28$ & $3.09$ & $\XbAeDslMs$ & $\XbAeDslVsPtwoZ$ & $\XbAeDslVsPtenZ$ & $5{,}144$ LLM calls \\
    \midrule
    Relay-BP (pkg) & published, stop 9 & $\XbRelayLer$ & --- & --- & $\OctRelayNineMs$ & $\XbRelayVsPtwoZ$ & $\XbRelayVsPtenZ$ & \texttt{relay-bp} package \\
    P5 Relay-BP & panel port, stop 9 & $\XbPfiveLer$ & $\OctAdjPFiveCostXPTwoD$ & $\OctAdjPFiveTimeXPTwoD$ & --- & $\XbPfiveVsPtwoZ$ & $\XbPfiveVsPtenZ$ & DSL port of Relay-BP \\
    Relay-BP (pkg) & published, stop 5 & $\XbRelaySfiveLer$ & --- & --- & $\OctRelayFiveMs$ & $\XbRelaySfiveVsPtwoZ$ & $\XbRelaySfiveVsPtenZ$ & \texttt{relay-bp} package \\
    P10 BP+AC & panel (faithful) & $\XbPtenLer$ & $\OctAdjPTenCostXPTwoD$ & $\OctAdjPTenTimeXPTwoD$ & $\SepAcMs$ & $\XbPtenVsPtwoZ$ & --- & 45 prod-sum sweeps, $\kappa{=}0.1$ \\
    P3 BP+OSD-CS & panel (\texttt{ldpc} pt.) & $\XbPthreeLdpcLer$ & $\OctAdjPThreeCostXPTwoD$ & --- & $\OctDampPThreeMs$ & $\XbPthreeLdpcVsPtwoZ$ & $\XbPthreeLdpcVsPtenZ$ & 30 serial sweeps, order 60 \\
    P3-d & panel (damped CS) & $4.22$ & --- & --- & $\OctDampPThreeDMs$ & $+28.8$ & $-21.1$ & 10 damped sweeps, order 60 \\
    P2 BP+OSD-0 & panel (\texttt{ldpc} pt.) & $\XbPtwoLdpcLer$ & $\OctAdjPTwoCostXPTwoD$ & --- & $\OctDampPTwoMs$ & $\XbPtwoLdpcVsPtwoZ$ & $\XbPtwoLdpcVsPtenZ$ & 30 serial prod-sum sweeps \\
    P2-d (ref.) & panel reference & $\XbPtwoLer$ & $\OctAdjPTwoDCostXPTwoD$ & $\OctAdjPTwoDTimeXPTwoD$ & $\OctDampPTwoDMs$ & --- & $\XbPtwoVsPtenZ$ & 10 damped min-sum sweeps \\
    \bottomrule
  \end{tabular}}
\end{table}

\begin{figure}[ht]
{\footnotesize
\begin{verbatim}
def solve_e083(state):   # Hybrid (LLM + evolution): 3.17% LER (split B), 36.2 ms/shot
    state = run_bp(state, 18, BpRule.PRODUCT_SUM, 1.0, 0.0, Schedule.PARALLEL, 0.625)
    state = reduce_marked_rows(state, True)
    state = enrich_blocks(state, 0.04, True)
    state = enumerate_patterns(state, 1, 16, Scope.OSD_CS)
    return vote_logical_classes(state)

def solve_c03(state):    # Fast-front, seeded: 4.17% LER (split B), 5.66 ms/shot
    state = run_bp(state, 9, BpRule.MIN_SUM, 0.7710, 0.0031, Schedule.PARALLEL, 0.3107)
    state = reduce_marked_rows(state, True)
    state = enrich_blocks(state, 0.0412, True)
    return enumerate_patterns(state, 1, 4000, Scope.BLOCK_VOTE)
\end{verbatim}
}
\caption{Two discovered decoders in the $\HypDslOps$-operation DSL, after dead-code pruning
(prediction-identical to the unpruned programs on $\NovConfirmShots$ validation shots).
\texttt{solve\_e083} (from the LLM-in-the-loop hybrid arm) combines ambiguity clustering's
marked-row reduction and narrow block enrichment ($\kappa = 0.04$) with a weight-$1$
combination-sweep search (\texttt{Scope.OSD\_CS}) and a logical-class majority vote, sitting
at call-multiset distance $2$ from P10 (BP+AC) and reducing its error rate by $7.6\%$
($z = +7.5$). \texttt{solve\_c03} replaces BP+AC's $45$-sweep product-sum front end and
weight-$2$ search with $9$ damped min-sum sweeps and a weight-$1$ block vote, cutting
errors by $\SepCThreeErrCutVsPTwoD$ relative to P2-d at $\SepCThreeTimeXPTwoD\times$ its time.}
\label{lst:qec-programs}
\end{figure}

\paragraph{Training-shot stratification.} A \emph{shot} is one sample of the
syndrome-extraction circuit's noisy measurement record together with the true logical
outcome; decoding it yields a correction that is either logically correct or erroneous. At
$p = 0.004$, most shots are decoded correctly by plain belief propagation and by every
candidate, so a uniform training draw spends most of its evaluation budget on shots that do
not separate competing post-processors. The campaigns reported in \S\ref{sec:qec} therefore
draw training shots using a two-stratum, importance-weighted sampler that over-samples the
stratum \emph{where plain BP fails}---the shots on which the post-processing stages
decide the outcome---while weighting each stratum by its population share so that the
estimated logical error rate remains unbiased. Because the stratum boundary is defined by
plain BP (a fixed panel member) on the training split alone, it is independent of the
candidate being evaluated. Conversely, stratifying on \emph{shots where the reference
decoder BP+OSD-0 fails} rewards candidates for exploiting the reference's specific failure
modes and inverted pairwise decoder rankings on nine of ten benchmarked pairs; that stratum
is therefore excluded. Every exam result is evaluated on an unweighted, uniform draw of the
exam split.

\end{document}